\documentclass[conference]{IEEEtran}
\IEEEoverridecommandlockouts
\usepackage{cite}
\usepackage{amsmath,amssymb,amsfonts}
\usepackage{algorithmic}
\usepackage{graphicx}
\usepackage{textcomp}
\usepackage{xcolor}
\usepackage{multicol}

\def\BibTeX{{\rm B\kern-.05em{\sc i\kern-.025em b}\kern-.08em
    T\kern-.1667em\lower.7ex\hbox{E}\kern-.125emX}}
\begin{document}

\title{Perlin Noise Improve Adversarial Robustness\\
	
	\thanks{This article is supported in part by the National Key R\&D Program of China under projects 2020YFB1006003 and 2020YFB1006004, the National Natural Science Foundation of China under projects 61862015, 61772150, 61862012 and 61962012, the Guangdong Key R\&D Program under project 2020B0101090002, the Guangxi Natural Science Foundation under grants 2018GXNSFDA281054, 2019GXNSFFA245015, 2019GXNSFGA245004 and AD19245048, and the Peng Cheng Laboratory Project of Guangdong Province PCL2018KP004.}
}

\author{\IEEEauthorblockN{Chengjun Tang}
	\IEEEauthorblockA{\textit{School of Computer Science and Information Security} \\
		\textit{Guilin University of Electronic Technology}\\
		Guilin, China \\
		0000-0003-4781-8662}
	\and
	\IEEEauthorblockN{Kun Zhang}
	\IEEEauthorblockA{\textit{ National Information Center} \\
		Beijing,
		China\\
		ndrczk@163.com}
	\and
	\IEEEauthorblockN{Chunfang Xing}
	\IEEEauthorblockA{\textit{Library of Guilin University of Electronic Technology} \\
		\textit{Guilin University of Electronic Technology}\\
		Guilin, China \\
		565108930@qq.com}
	\and
	\IEEEauthorblockN{Yong Ding}
	\IEEEauthorblockA{\textit{School of Computer Science and Information Security} \\
	\textit{Guilin University of Electronic Technology}\\
	Guilin, China \\
	0000-0002-3571-7576}
	\and
	\IEEEauthorblockN{Zengmin Xu* }
	\IEEEauthorblockA{\textit{School of Mathematics and Computing
			Science} \\
		\textit{Guilin University of Electronic Technology}\\
		Guilin, China \\
		zengminxu@gmail.com}
}

\maketitle

\begin{abstract}
	Adversarial examples are some special input that can perturb the output of a deep neural network, in order to make produce intentional errors in the learning algorithms in the production environment. Most of the present methods for generating adversarial examples require gradient information. Even universal perturbations that are not relevant to the generative model rely to some extent on gradient information. Procedural noise adversarial examples is a new way of adversarial example generation, which uses computer graphics noise to generate universal adversarial perturbations quickly while not relying on gradient information. Combined with the defensive idea of adversarial training, we use Perlin noise to train the neural network to obtain a model that can defend against procedural noise adversarial examples. In combination with the use of model fine-tuning methods based on pre-trained models, we obtain faster training as well as higher accuracy. Our study shows that procedural noise adversarial examples are defensible, but why procedural noise can generate adversarial examples and how to defend against other kinds of procedural noise adversarial examples that may emerge in the future remain to be investigated.
\end{abstract}


\section{Introduction}
Convolutional neural networks are widely used in the domain of computer vision, such as image recognition\cite{ILSVRC15} due to their powerful representation capabilities. However, as image features become more complex, more extensive scale, in particular, deeper layers of neural networks are in demand. Unfortunately, neural networks seem to have a bug of being vulnerable to adversarial examples, which uses almost invisible perturbations to make the neural network change its output. \cite{FGSM} Early methods of generating adversarial samples are usually based on the gradient information of neural networks, such as PGD and FGSM. \cite{FGSM, PGD} Although these adversarial example can be successfully attacked with small perturbations, these adversarial samples usually cannot be migrated to other models nor applied to physical objects. Procedural noise functions can generate adversarial examples that can successfully perturb almost all model. \cite{PerlinAE} This technique allows the application of computer graphics techniques to the neural networks of computer vision. There is no good defense method for procedural noise adversarial example. In Co et al.'s work \cite{PerlinAE} , they provide a preliminary defense, denoising, which cannot and even will not be able to successfully defend against this attack. We can give a theoretical proof of it. Therefore, this paper proposes a method similar to noise training and adversarial training, which can effectively defend against procedural noise adversarial example. However, this method cannot provide an effective defense against gradient-based methods such as FGSM. We therefore propose a method based on Perlin noise masking to try to obfuscate the adversarial input. The defense rate of this method is too low and therefore not practical enough, but it is still relevant for discussion. 

In this paper, we redefine the procedural noise adversarial examples of our improved version. We use transfer learning and model fine-tuning in our experiments. Then we explain our Perlin noise training model for the defence of procedural noise adversarial example. Finally we give the solution based on stacked and achie accuracy that matches the ensemble adversarial training \cite{EnsAdvT}.

\section{Related works}
\label{sec:disc}

Yang et al. and Huang et al. use some computer graphics technologies to simulate real scene, thus enabling physical attacks. \cite{PhyCG, PhyCG2} Perlin noise is also an computer graphics technology and Perlin noise also can be used for physical attacks, which inspired us to use more computer graphics methods to enhance the task of computer vision. Cihang et al.'s work show that input diversity could improve adversarial transferability. \cite{diversity} Our Perlin noise mask can improve input diversity for defense. What input diversity means for neural networks is still an element worth investigating. We use a model fine-tuning technique to solve the problem that Perlin noise training from scratch fails to converge. Chen et al. discusses a method based on model fine-tuning that can simultaneously improve the accuracy and robustness of the model. \cite{fineadv} Other data augmentation methods that lead to difficult model convergence may be able to make progress in combination with model fine-tuning. AutoAugment manifests a great effect on the accuracy improvement of image classification neural networks. \cite{AutoAugment} Our network trained with only Perlin noise augmentation also achieves an accuracy close to that of EfficentNet using many training techniques, including AutoAugment. \cite{EfficientNets}

\subsection{Noise training and Perlin noise}

Most related to our work in noise training is RazorNet introduced by Taheri et al.\cite{Taheri2019}. RazorNet uses Perlin noise in the training of neural networks to improve accuracy and robust, but they did not discuss why Perlin noise was used or the difference between Perlin noise and other noise. Perlin noise was first proposed by Ken Perlin in 1985 as an image synthesizer rather than a type of noise\cite{Perlin1985}. Perlin noise was used as a texture generator in games to simulate objects such as smoke.\cite{PerlinSmoke} Bae et al. used Perlin noise in data augmentation to solve small data samples of HRCT (high-resolution computed tomography) images.\cite{PerlinDA} However, they have no more in-depth experimental experiment on how to mix the augmented data with the original data and how Perlin noise interacts with the original image. The work of Lecuyer et al.\cite{DP} is most well known, where they inserted additive noise into neural networks based on differential privacy theory to obtain models with certified robustness. The effect of additive Gaussian noise in the robustness of neural networks was well investigated by Li et al.\cite{AddG}

\subsection{Adversarial attack and adversarial training}

Co et al. black-box adversarial example attacks and gives a defense method based on denoise, which did not work so well.\cite{PerlinAE} They also did not mention any experiments on low-resolution datasets. And according to our experiments, procedural noise adversarial examples behave very differently on low-resolution datasets. Specifically, it does not have a similar perturbation universality as on the high-resolution dataset, which requires further research. Free Adversarial Training and \cite{AdvFree} Fast Adversarial Training \cite{AdvFast}, YOPO \cite{YOPO} improve speed of adversarial training. Our approach uses Perlin noise, a universal adversarial perturbation, for training and masking, which is faster than them. But it didn't achieve the defensive effect to match. Other universal adversarial perturbations should perhaps be equally noted for application in adversarial training. The work of Zhang et al. shows that the adversarially trained network recognizes shapes more than textures, which explains the resistance of the adversarially trained model to Perlin noise attacks, because Perlin noise is a texture generation method. \cite{InterpAdvModel} The success of our approach encourages the use of other texture generation methods to augment the data.

\section{Procedural noise adversarial examples}
\label{section:noiseadv}
Co et al. introduced procedural noise functions as an intuitive and computationally efficient approach to generate adversarial examples in black-box settings.\cite{PerlinAE} As a kind of lattice noise, Perlin noise determines noise at a point in space by computing a pseudo-random gradient at each of the eight nearest vertices on the integer cubic lattice and then doing a splined interpolation. \cite{Perlin1985, Perlin2002} 

As Bishop et al.'s work, adding random noise with zero mean and uncorrelated between different inputs is equivalent to Tikhonov Regularization. \cite{Bishop1995} 
After adding Perlin noise $\boldsymbol{\xi}$
%
and expand the network function as a Taylor series powers of $\xi$ we can get: 

\resizebox{0.45\textwidth}{!}{
	$
	y_{k}(\mathbf{x}+\boldsymbol{\xi})=y_{k}(\mathbf{x})+\left.\sum_{i} \xi_{i} \frac{\partial y_{k}}{\partial x_{i}}\right|_{\boldsymbol{\xi}=0}+\left.\frac{1}{2} \sum_{i} \sum_{j} \xi_{i} \xi_{j} \frac{\partial^{2} y_{k}}{\partial x_{i} \partial x_{j}}\right|_{\boldsymbol{\xi}=0}+\mathcal{O}\left(\boldsymbol{\xi}^{3}\right)
	$
}

Different from Bishop et al.'s, although we add Perlin noise $\boldsymbol{\xi}$ with zero mean and to be uncorrelated between different inputs, we have no $\mathcal{O}\left(\boldsymbol{\xi}^{3}\right)=0$.
So adding randomly Perlin Noise is different from noise training with Gauss Noise.

\begin{figure}
	\centering
	\begin{multicols}{2}
		\includegraphics[width=0.5\linewidth]{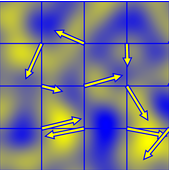}
		\includegraphics[width=0.5\linewidth]{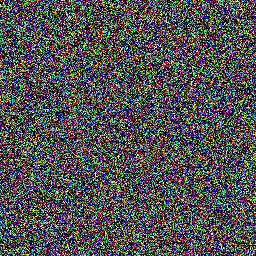}	
	\end{multicols}
	
	\caption{Perlin Noise and Gauss Noise}
	\label{fig:perlinn}
\end{figure}

\section{Methods}
\label{sec:methods}

In this section, we elaborate our method for improving adversarial robustness. As described in Section~\ref{sec:advnoi}, Perlin noise is used to generate adversarial examples or augment data stochastically. We introduce an improved method that adds transform operation to Co et al.'s\cite{PerlinAE}. Different ratios of augmented data is tried to observe the performance. In robust metrics, attacks are considered, including Fast Gradient Sign Method (FGSM)\cite{FGSM}, Projected Gradient Descent (PGD)\cite{PGD}, Carlini and Wagner $L_2$ and $L_\infty$ Attack\cite{Carlini2017}. 

\subsection{Training methods}

\paragraph{Transfer learning and fine-tuning}
Transfer learning, in other words, pretraining, is another exciting paradigm to improve robustness and save much training time. \cite{Weiss2016, Shao2015} In this work, pretraining uses a network on the big dataset  ImageNet and then using its weights as the initial weights in subsequent noise training and adversarial training. Adding noises or adversarial examples at the beginning can easily make training harder to converge and have a more considerable bias, but smaller variance, which means better robustness most of the time. \cite{Zhou2019} Transfer learning can reduce the time training a neural network from scratch on ImageNet dataset. Comparing a transfer learning trained model with a model trained from scratch can explain why adversarial examples appear and how adversarial training affects the neural network.

In this work, we define the neural network as a function stacked by some functions $H_i$ called layers, and each layer has a set of trainable parameters, $W_i$. i.e., $H_n(H_{n-1}(...), W_n)$. (In fact, there may some connections cross layers, but we still describe it this way for simplicity. ) For the outermost layer $H_n$ we call it the top layer. In our solution, we freeze the weights other than the top layer and thus fine-tuning.


\paragraph{Adversarial training and noise training} \label{sec:advnoi} As described in \ref{section:noiseadv}, Perlin Noise can be used as a procedural noise or a method to generate adversarial samples. Using Bayesian optimization to optimize the Perlin Noise function parameters can be used to generate adversarial examples for adversarial training. Using random parameters to generate Perlin noise, which is added into the train set, can be regarded as an implementation of noise training. Comparing the results of adversarial training and noise training using Perlin noise can interpret the difference between them.

\subsection{Robustness metrics}

How to measure the robustness of the model is a topic worth discussing. Local loss sensitivity is estimated through the gradients of the prediction at points in the dataset. \cite{Arplt2017} Weng et al.'s method CLEVER can be better quantified than loss sensitivity to compare different models' robustness. However, local methods may have specific limitations. Using attack methods to measure the accuracy under attack can also be used as a robust metric. So Fast Gradient Sign Method (FGSM)\cite{FGSM}, Projected Gradient Descent (PGD)\cite{PGD}, Carlini and Wagner $L_2$ and $L_\infty$ Attack\cite{Carlini2017} are selected.

\section{Experiments}
\label{sec:exp}
In this section, we describe our experiment settings firstly. Due to the differences in the performance of procedural noise adversarial examples on different resolution data sets, we divide the experimental discussion into two parts: high-resolution data sets and low-resolution data sets. In each part, every models' robustness is evaluated by attacks, robustness metrics and robustness datasets.

\subsection{Experiments Settings}
\paragraph{High Resolution Dataset}
We conduct experiments on ImageNet 2012 ILSVRC challenge prediction task since it has been considered one of the most heavily benchmarked datasets in computer vision and there are some sub datasets such as ImageNet-A which could be used to evacuate robustness.

\paragraph{Low Resolution Dataset}
Perlin noise looks very different at low resolution than at high resolution. So CIFAR10 with 32x32 low resolution has been considered one of the most benchmarked datasets in computer vision image classify. CIFAR10 has 10 classes containing 6000 images each. There are 5000 training images and 1000 testing images per class.

\paragraph{Architecture}
We use EfficientNets \cite{EfficientNets} as the SOAT model of image classify. And for more interpretability, we also use ResNets \cite{ResNets} as one of the most widely used models with lesser confusing hyperparameter adjustment. The ResNet50 and EfficientNet-B0 were chosen for the main experiment due to the computing capacity limitations of the experimental equipment. More trainable parameters are of little relevance to our conclusions.

\paragraph{Training Settings}
For low resolution dataset, we use batch size of 128 by default because larger batch size even makes models slower to converge. For high resolution dataset, we use batch size of 32 to pre, because of the memory limit of experiment equipment. We use NAdam optimizer \cite{NAadm} for all our experiments but adjust learning rate and use decay at different experiments.

\paragraph{Noise or Adversarial Training}
For noise training we add a random noise to each image sample. Generating procedural noise adversarial examples requires a lot of computational capacity, so a cache is used. If a noise can effectively perturb the model's output, then this noise is cached and will be firstly tried in subsequent searches.

\subsection{Experiments on High Resolution Dataset}

\begin{figure}
	\centering
	\includegraphics[width=0.7\linewidth]{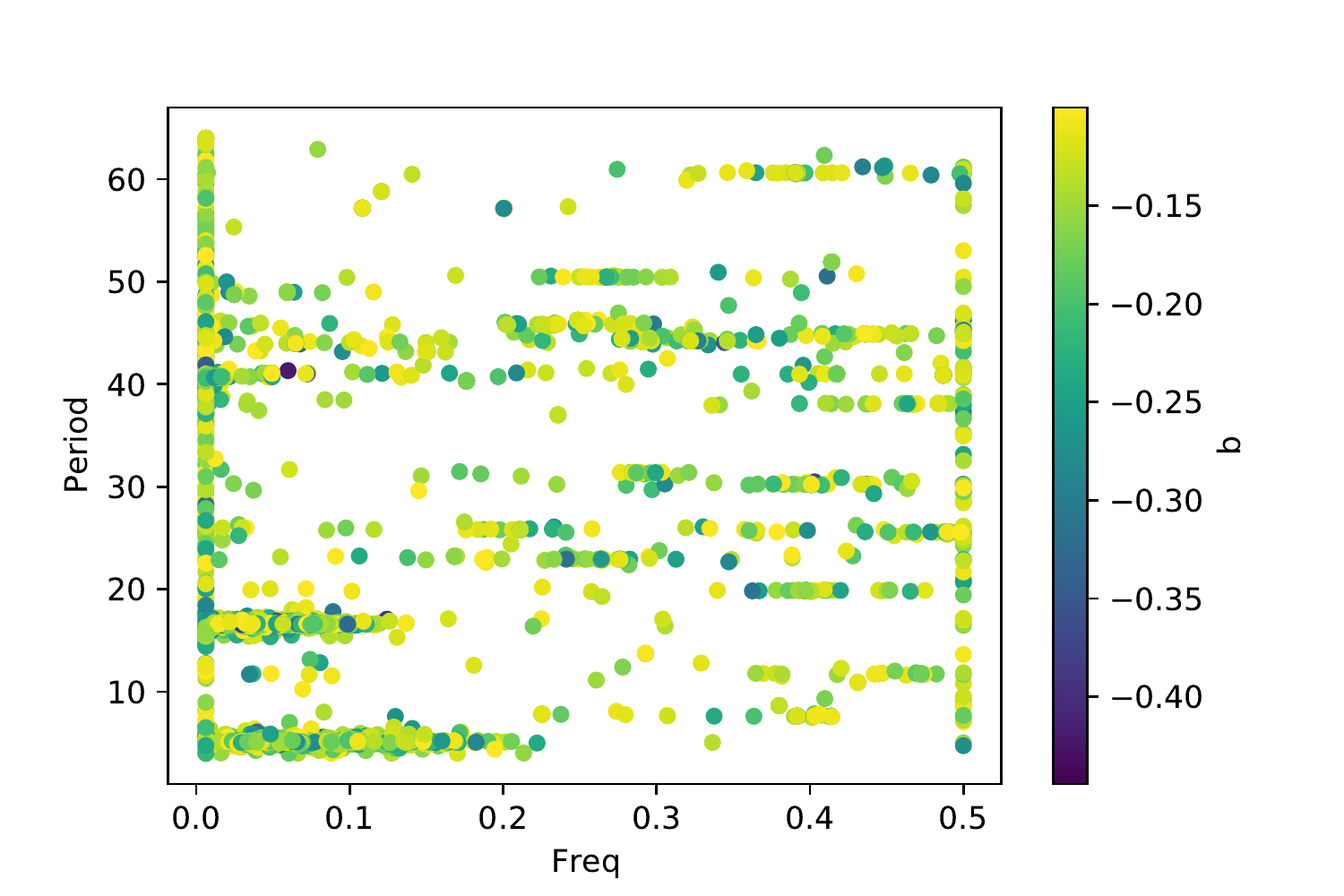}
	\caption{Scatter plot of parameters for successful attacks}
	\label{fig:boptparam}
\end{figure}

\paragraph{Pre-experiments on Parameters' Bayesian Optimization.}
In order to make a preliminary comparison between the Bayesian optimization method in Co. et al.\cite{PerlinAE} and the direct random generation of procedural noise, the two methods were used to generate perturbed images, and finally a karate-test was used to compare the misclassification of the samples generated by the two methods.
\begin{table}
	\centering
	\caption{Pre-experiment result}
	\begin{tabular}{|c|c|c|c|}
		\hline
		& Random & Co et.al &  \\
		\hline
		Correct & 3390 & 2063 & 5453 \\
		\hline
		Misclassification & 6610 & 7937 & 14547 \\
		\hline
		& 10000 & 10000 & 20000 \\
		\hline
	\end{tabular}
\end{table}
$$H_0: \text{Additional Bayesian optimization has no effect}$$
$\chi^2= 443.98$,$P(\chi^2)=0.00$, reject $H_0$. Additional Bayesian optimization is effective. We then plot a scatter of the Bayesian optimized parameters which is shown in figure~\ref{fig:boptparam}. Observation of the scatter plot shows that Freq is distributed centrally around the levels of 26, 17, and 7. Meanwhile Period is concentrated at the level of 2.5 and 40 on both sides. Further labeling estimates based on this phenomenon were imposed.
\paragraph{Pre-experiments on Label Search.}
According to the distribution of the attack success parameters, they are divided into three components, which are estimated separately using normal distribution. Randomly generate noise samples according to each of the three estimated normal distributions. The label shifts due to perturbation of the input by the noise generated by the three distributions are counted separately. We found that the predicted labels of one group of distributions were concentrated in doormat and fountain, rather than brain coral and shower curtain, which were similar to the other distributions.
\begin{table}
	\centering
	\caption{Pre-experiment label result}
	\begin{tabular}{|c|c|c|c|}
		\hline
		\multicolumn{2}{|c|}{Freq=40 Period=2.5} & \multicolumn{2}{|c|}{Freq=32 Period=60}   \\
		\hline
		Count & Label & Count & Label \\
		\hline
		370 & Doormat & 1313 & Brain coral \\
		\hline
		238 & Fountain & 453 & Spider web \\
		\hline
		216 & Television & 407 & Shower curtain \\
		\hline
	\end{tabular}
\end{table}

\begin{figure}
	\centering
	\includegraphics[width=0.6\linewidth]{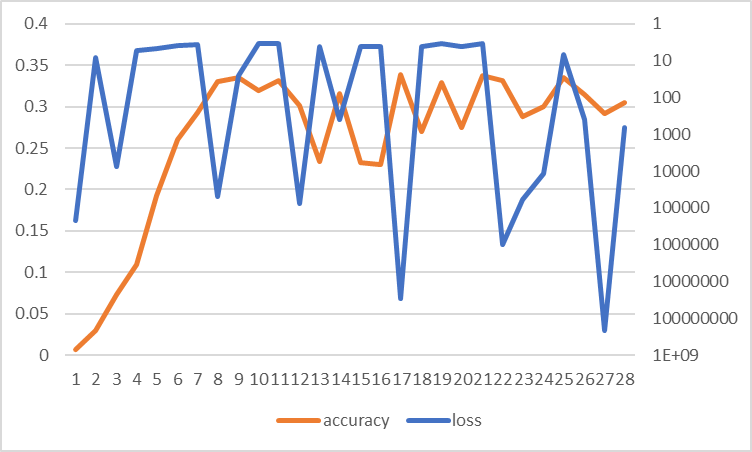}
	\caption{Validation set accuracy and loss function values for noise training from scratch}
	\label{fig:noiefn}
\end{figure}

\paragraph{Pre-experiments on Pretraining model.}
We first train the EfficentNetB0 model with Perlin noise augmentation directly from scratch, but the model accuracy stays below 60\%. This is not only true for natural data, but also for augmented data. This leads us to believe that adding Perlin noise for training makes the model difficult to converge. Our attempts to reduce the proportion of data containing the augmented data and our attempts to train using the adversarial examples obtained from the Bayesian optimization search described previously did not work.
Since training Imagenet \cite{ILSVRC15} data from scratch is difficult to converge (which is shown in Figure~\ref{fig:noiefn}), it is an exploration of how to use with the training model for fine-tuning. We use three types of fine-tuning, initializing the top layer randomly and then fine-tuning only the top layer, fine-tuning only the top layer, and fine-tuning the whole model. We use the NAdam optimizer \cite{NAadm} to fine tune with a learning rate of 0.0001. Finally, we find that full-layer fine-tuning is more accurate on both clean and noise augmented data. 

\begin{table}
	\centering
	\caption{Fine-tuning strategy experiments}
	\begin{tabular}{|c|c|c|c|c|c|c|}
		\hline
		Strategy & \multicolumn{2}{c|}{Full layer} & \multicolumn{2}{c|}{Top layer} &   \multicolumn{2}{c|}{Top layer random init}   \\
		\hline
		Epoch & Clean & Noise & Clean & Noise & Clean & Noise \\
		\hline
		1 & 60.89 & 66.23 & 49.30 & 46.44 & 54.80 & 46.16 \\
		\hline
		2 & 70.96 & 67.59 & 46.45 & 46.45 & 53.08 & 47.83 \\
		\hline
	\end{tabular}
	\label{tab:finetune}
\end{table}

\paragraph{Comparison Experiment: Adversarial Training vs. Noise Training.}
We now evaluate our Perlin Noise Training and Adversarial Training strategy described in Section~\ref{sec:advnoi}. Repeating our strategy, we use randomly generated Perlin noise and the disturbance sample obtained from Bayesian optimization to training models. The parameters of the randomly generated Perlin noise are determined according to the distributions obtained from the search in this section. The frequencies were set to a normal distribution with a mean of 40, while the periods' were set to 2.5 and 60, respectively. It took too much time to search for the perturbation that caused the largest drop in confidence in the correct classification for each example (we spent about 80 hours with 20 initialization queries and 20 maximum iterations for the 50,000 validation set of Imagenet \cite{ILSVRC15}) so we stopped the perturbation once we got a prediction label that would cause the model to output an error and cached the data. Most of the data were perturbed with successful examples during the initialization query. According to such a strategy, the training sets for adversarial and noisy training should differ only by a small number of samples. But in fact, the accuracy of the model trained against the sample is always lower than that of the noise trained. This is the case in both full model and top layer fine-tuning.
\paragraph{Solution with Perlin noise mask}
\label{sec:fin}
\begin{figure}
	\centering
	\includegraphics[width=0.3\linewidth]{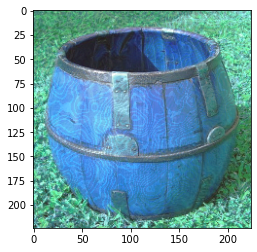}
	\caption{Perlin noise masked FGSM adversarial image.}
	\label{fig:masked}
\end{figure}

Based on the above experimental results, we choose the model after 2 epochs' full-layer fine-tuning training as our final adopted model. As shown in Table~\ref{tab:finetune}, since our model is robust to procedural noise adversarial examples, perturbing the input of the model with procedural noise may be able to perturb the adversarial examples generated by other attack methods so that they can be defended against. Let the Perlin noise trained neural model as $H_{\mathrm{per}}$, and Perlin noise mask as $G_{\mathrm{per}}$, raw image as $I$, then we have $H_{\mathrm{per}}(\min(I+G_{\mathrm{per}}, 1))$ as the solution's output, which is shown in Figure\ref{fig:masked}. We finally tested its accuracy under some attack methods, as shown in Table~\ref{tab:fin}. The model after only fine-tuning can only defend against a single-step FGSM attack, while it can get 22.74\% accuracy under PGD attack after procedural noise masking to process the adversarial input. The effect of this method is relatively insignificant compared to Ensemble Adversarial Training \cite{EnsAdvT}. However, since the main purpose of this method is to defend against procedural noise examples, this result is acceptable.

\begin{table}
	\centering
	\caption{Perlin noise mask model' accuracy over FGSM, PGD and Perlin noise attack, compare with Ensemble Adversarial Training \cite{EnsAdvT}}
	\begin{tabular}{|c|c|c|c|c|}
		\hline
		Source & FGSM & PGD & Noise & Target \\
		\hline
		EfNB0 & 0.00 & 0.00 & 34.92 &  EfNB0 \\
		\hline
		EfNB0 & 17.40 & 12.04 & - & EfNB0$_{adv}$ \\
		\hline
		EfNB0$_{adv}$ & 14.80 & 0.00 & 59.54 & EfNB0$_{adv}$ \\
		\hline
		EfNB0$_{adv}$ & 27.80 & 22.74 & - & EfNB0$_{mask}$ \\
		\hline
	\end{tabular}
	\label{tab:fin}
\end{table}

\paragraph{Speed Test}
In terms of adversarial training speed optimization, both Free Adversarial Training and \cite{AdvFree} Fast Adversarial Training \cite{AdvFast} improvements inevitably require the evaluation of the FGSM \cite{FGSM} at least once. Therefore, we compare the time to compute the FGSM algorithm once by the backpropagation algorithm with the time we take to generate a random Perlin noise. The Perlin noise algorithm in our experiments is an CPU implementation. We therefore tested the time to calculate FGSM once on the CPU and once on the GPU separately for comparison. As shown in Table~\ref{tab:speed}, it took an average of 115ms to generate the FGSM examples on the GPU and 52ms to generate the Perlin noise examples on the CPU, which means that the enhancement using Perlin noise consumes less than half of the time to generate the FGSM examples. Since we are able to generate Perlin noise examples on the CPU in parallel, our training strategy can achieve the same speed as the non-adversarial training approach. This means that procedural noise adversarial examples are no longer available as a UAP (universal adversarial perturbation) on low-resolution datasets like CIFAR10.

\begin{table}
	\centering
	\caption{Speed test result}
	\begin{tabular}{|c|c|c|c|}
		\hline
		& Perlin & FGSM GPU & FGSM CPU \\
		\hline
		Mean & 52.3 & 115 & 284 \\
		\hline
		Std. &  5.87 & 4.66 & 5.89 \\
		\hline
	\end{tabular}
	\label{tab:speed}
\end{table}

\subsection{Experiments on Low Resolution Dataset}

\begin{figure}
	\centering
	\includegraphics[width=0.3\linewidth]{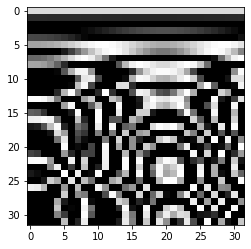}
	\caption{Low resolution Perlin Noise}
	\label{fig:lowres}
\end{figure}

There are no experiments on datasets other than ImageNet\cite{ILSVRC15} in the paper of Co et al.\cite{PerlinAE}. And Perlin Noise looks different in low resolution, which is shown in Figure\ref{fig:lowres}.  Co et al. proposed a sine color map in order to generate more distinguishable textures, while the low resolution corrupts it. First we trained a ResNet50\cite{ResNets} network based on the CIFAR10 dataset\cite{CIFAR}. The network can achieve 80\% accuracy on the clean CIFAR10 dataset. We then attack the entire test set using the Bayesian optimization method mentioned in Section~\ref{section:noiseadv}. This eventually allowed 9,997 out of 10,000 samples to be successfully perturbed for output. Moreover, the success rate of the attack for the original model to output the results correctly is 100\%. However, our attention was drawn because the randomly generated perturbation using the distribution obtained based on Bayesian optimization had only a 29.16\% success rate. This means that procedural noise adversarial examples are no longer available as a universal adversarial perturbation on low-resolution datasets like CIFAR10.  
Repeating our defense method, which is mentioned in Section~\ref{sec:fin}, on the low-resolution dataset. It is found that noise masks reduce the accuracy rather than prevent model from adversarial attack. 

\subsection{Stacked Solution}
In Table\ref{tab:fin} we find our Perlin noise training model and Perlin noise mask solution have low accuracy over FGSM or PGD attack.  In order to solve this problem we apply our transfer learning method on Kurakin et al.\cite{EnsTF}'s Inception v3 on ensemble of 4 models model which based on Tram{\`{e}}r et al.\cite{EnsAdvT}'s method. And we gain about seventy percent accuracy over FGSM\cite{FGSM} The accuracy of the model trained by this method still has a certain gap compared to Kurakin et al.\cite{EnsTF}'s model. So we train a Perlin noise detection model. Then we use the stacked\cite{stackedBreiman, stackedWolpert} method to train a decision tree model by combining the outputs of Kurakin et al.\cite{EnsTF}'s model and  Perlin noise detection model, and finally we can get the accuracy matching it, which is shown in Table\ref{tab:stack}. 

\begin{table}
	\centering
	\caption{Perlin noise mask result}
	\begin{tabular}{|c|c|c|c|}
		\hline
		Source & FGSM & PGD & Noise \\
		\hline
		Inv3E4 & 76.6 & 73.5 & 20.6 \\
		\hline
		Inv3E4$_{noi}$ & 75.9 & 72.9 & 73.6 \\
		\hline
		Inv3E4$_{stacked}$ & 76.6 & 73.5 & 73.6 \\
		\hline
	\end{tabular}
	\label{tab:stack}
\end{table}

\section{Conclusion}
\label{sec:con}
Procedural noise adversarial examples lack effective defense methods. In this paper, a method between noise training and adversarial training is proposed for its defense. Model fine-tuning training is performed after data enhancement using random Perlin noise. Not similar to the model trained by ensemble adversarial training \cite{EnsAdvT} has can defense against Perlin noise adversarial examples in some extent, this approach does not directly defend against gradient-based attacks such as PGD\cite{PGD}. Therefore, we propose a masking method using Perlin noise to confuse the gradient-based adversarial perturbations, which achieved some results but it is insignificant when compared to ensemble adversarial training. In addition, we found that Perlin noise behaves differently at high and low resolution. At high resolution it is a universal adversarial perturbation while at low resolution it is not. We think it is an issue to explore whether Perlin noise enhancement can be used to improve accuracy. 

\begin{figure}
	\centering
	\includegraphics[width=0.7\linewidth]{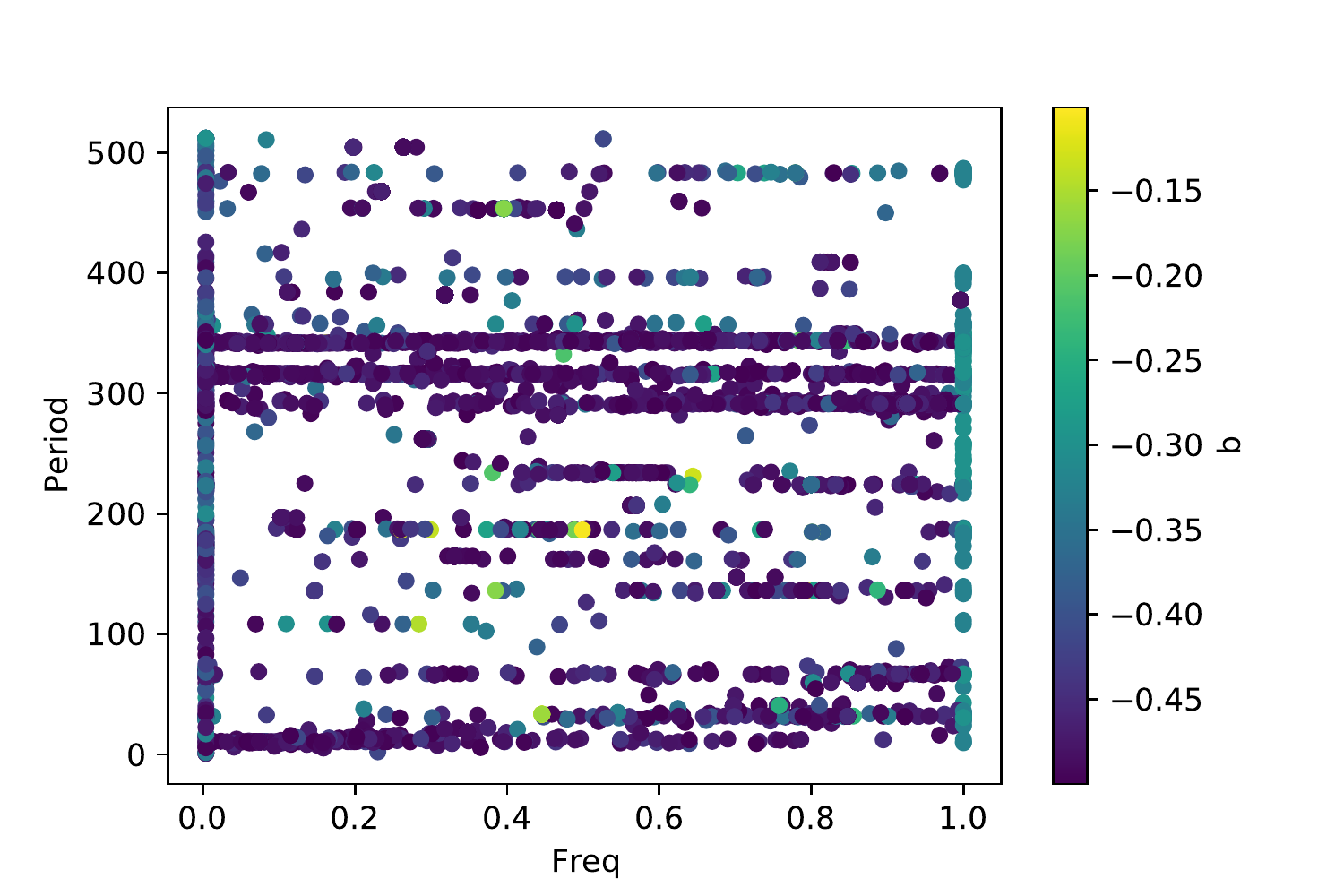}
	\caption{Scatter plot of success attack parameters for CIFAR10}
	\label{fig:figc}
\end{figure}

\bibliography{document}

\end{document}